\documentclass[]{spie}  

\usepackage{amsmath,amsfonts,amssymb}
\usepackage{graphicx}
\usepackage[colorlinks=true, allcolors=blue]{hyperref}
\usepackage{tabularx}

\usepackage{xcolor}
\newcounter{rainbowcounter}

\newcommand{\rainbowhelper}[1]{%
  \ifx#1\relax
  \else
    \rainbowletter{#1}%
    \expandafter\rainbowhelper
  \fi
}
\newcommand{\rainbowletter}[1]{%
  \ifcase\value{rainbowcounter}%
    \textcolor{red}{#1}
  \or
    \textcolor{orange}{#1}
  \or
    \textcolor{darkorange}{#1}
  \or
    \textcolor{green}{#1}
  \or
    \textcolor{blue}{#1}
  \or
    \textcolor{violet}{#1}
  \else
    \textcolor{red}{#1}
  \fi
  \stepcounter{rainbowcounter}%
  \ifnum\value{rainbowcounter}=6 \setcounter{rainbowcounter}{0}\fi
}

\definecolor{darkorange}{rgb}{0.8,0.4,0.0}

\newcommand{\red}{\color[rgb]{0.6,0.0,0.0}}

\newcommand{\blue}{\color[rgb]{0.0,0.6,0.6}}

\newcommand{\ari}[1]{{[\bf \blue {Ari: } #1}]}

\newcommand{\peter}[1]{{[\bf \red {Peter: } #1}]}

\title{A Framework for the Assurance of AI-Enabled Systems}

\author[b]{Ariel S. Kapusta}
\author[a]{David Jin}
\author[a]{Peter M. Teague}
\author[a]{Robert A. Houston}
\author[a]{Jonathan B. Elliott}
\author[a]{Grace Y. Park}
\author[c]{Shelby S. Holdren}

\affil[a]{Office of the Chief Digital  and Artificial Intelligence Officer, U.S. Department of Defense, 9010 Defense Pentagon, Washington, D.C., USA}
\affil[b]{The MITRE Corporation, 7515 Colshire Drive, McLean, VA, USA}
\affil[c]{John Hopkins University Applied Physics Laboratory, 11100 Johns Hopkins Rd, Laurel, MD, USA}

\authorinfo{
Further author information:\\
(Send correspondence to David Jin)\\
David Jin: E-mail: {\tt\small david.jin5.civ@mail.mil}\\
Ariel S. Kapusta: E-mail: {\tt\small akapusta@mitre.org}\\
Cleared for Open Publication on Mar 11, 2025 by the U.S. DOD Office of Prepublication and Security Review, Case Number 25-T-1410.\\
Approved for Public Release, Distribution Unlimited. 
MITRE Public Release Case Number 25-0757\\
}

\begin{document} 
\maketitle

\begin{abstract}

The United States Department of Defense (DOD) looks to accelerate the development and deployment of AI capabilities across a wide spectrum of defense applications to maintain strategic advantages. However, many common features of AI algorithms that make them powerful, such as capacity for learning, large-scale data ingestion, and problem-solving, raise new technical, security, and ethical challenges.
These challenges may hinder adoption due to uncertainty in development, testing, assurance, processes, and requirements. 
Trustworthiness through assurance is essential to achieve the expected value from AI.

This paper proposes a claims-based framework for risk management and assurance of AI systems that addresses the competing needs for faster deployment, successful adoption, and rigorous evaluation. This framework supports programs across all acquisition pathways provide grounds for sufficient confidence that an AI-enabled system (AIES) meets its intended mission goals without introducing unacceptable risks throughout its lifecycle. 
The paper's contributions are a framework process for AI assurance, a set of relevant definitions to enable constructive conversations on the topic of AI assurance, and a discussion of important considerations in AI assurance. The framework aims to provide the DOD a robust yet efficient mechanism for swiftly fielding effective AI capabilities without overlooking critical risks or undermining stakeholder trust.

\end{abstract}


\newpage

\section{INTRODUCTION}
\label{sec:intro}  




The United States (U.s.) Department of Defense (DOD) recognizes the transformative potential of artificial intelligence (AI) in augmenting mission-critical systems for improved decision support, enhanced situational awareness, and autonomous operations. From data analytics for intelligence or logistics to sophisticated autonomous platforms in land, air, or maritime environments, AI-enabled systems (AIES) can offer improved speed and precision in addressing fast-evolving threats. 

The potential for AI reaches across a wide spectrum of defense applications. Some AIES focus on analytics and decision-support, synthesizing information from disparate data streams and generating insights to guide human decision-makers. Others involve more advanced autonomy with humans in-the-loop, on-the-loop, or out-of-the-loop. AI functions may drive unmanned platforms in air, land, or maritime domains, granting a capability to adapt to changing mission parameters without human intervention at each step. In many cases, the AI components may operate in environments where conditions change rapidly, inputs may deviate from training data, and unexpected scenarios may arise. Although it is unlikely for a passenger jet to land on a highway, autonomous cars are expected to avoid collisions, even in such rare events.


However, many of the same qualities that make AI so powerful—its capacity for learning, large-scale data ingestion, and problem-solving—raise new technical, security, and ethical challenges. AIES may experience drift, such as from input data drift, sensor degradation, actuation failures, or material fatigue. Some AI algorithms are designed to learn over time (either online or in increments), which is, functionally, internal drift in the AIES. Given these complexities, many traditional T\&E, safety, and cybersecurity processes—originally developed around deterministic software or hardware systems—can prove insufficient. Defense organizations accustomed to validating deterministic, conventional software systems may find themselves unprepared for the complexity, variability, vulnerabilities, uncertainties, brittleness, and opacity of many AIES.
Despite these challenges, to remain competitive the DOD must field AI-based solutions more quickly than has historically been possible. Many potential AI-based solutions have seen slow adoption, due to uncertainty in development, testing, assurance, acquisitions, processes, and requirements. 

This paper introduces a framework for risk management and assurance of AI systems that addresses the twin imperatives of speed and rigor across the entire lifecycle of the system. For consistency in understanding, we provide terminology definitions in Section \ref{sec:definitions}. By centering on explicit “assurance claims,” the framework provides a systematic way to capture all relevant evidence for a given AI-enabled system, from training data provenance to cybersecurity test reports and ethical compliance artifacts. In doing so, it brings together several often-stovepiped perspectives: safety, security, legal, and performance. 

The framework's process requires the creation of explicit arguments supported by evidence which ensure risks are understood and managed as required. The framework requires the creation of specific artifacts, including an assurance plan that defines how assurance is maintained over the system's lifecycle. These artifacts also provide information to stakeholders, supporting confidence in the AI system by those who must use, interact with, or be subjected to the system. 
With clear steps, a modular and iterative approach, tangible completion criteria, and explicit approval gates, this framework enables a faster path for developers to get new products into the hands of users. The clarity and consistency in requirements to achieve assurance are the primary means of acceleration provided by the framework. Future work to provide templates, guides, and examples that fit within the framework will further enable acceleration of development and adoption efforts. For example, a database of hazards common to AIES could increase consistency, quality, and speed of hazard identification and risk assessment activities.

\begin{figure}[t]
    \centering
    \includegraphics[width=\textwidth]{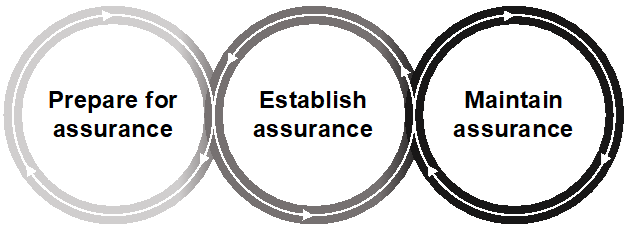}
    \caption{Phases of the AI assurance process}
    \label{fig:phases}
    \vspace{-0pt}
\end{figure}

The AI assurance process extends across the lifecycle of an AI system, from system development, through system fielding, and throughout system operations. Figure \ref{fig:phases} illustrates the three phases of the AI assurance process: prepare for assurance, establish assurance, and maintain assurance

This paper's key contributions are as follows:

\begin{itemize}
    \item A framework process for AI assurance. 
    \item Relevant definitions that may enable constructive conversations on the topic of AI assurance.
    \item Discussion of important considerations in AI assurance.
\end{itemize}

\section{Definitions}\label{sec:definitions}

For consistency in understanding, we provide the following terminology definitions that are used in this paper.

\begin{tabularx}{16cm}{>{\raggedright\arraybackslash}p{0.8cm} >{\raggedright\arraybackslash}p{3.0cm} X}
D-1 & \textbf{AI System} & An engineered or machine-based system that
can, for a given set of objectives, generate outputs such as predictions, recommendations,
or decisions influencing real or virtual environments. AI systems are designed
to operate with varying levels of autonomy. [Source NIST AI RMF\cite{nist-ai-rmf}]
\end{tabularx}

\begin{tabularx}{16cm}{>{\raggedright\arraybackslash}p{0.8cm} >{\raggedright\arraybackslash}p{3.0cm} X}
D-2 & \textbf{AI Assurance Process} & The process of managing risk, ensuring effectiveness, and leveraging the DOD AI ethical principles throughout the lifecycle of an AI system (D-1) to establish and maintain its assurance (D-3).
\end{tabularx}

\begin{tabularx}{16cm}{>{\raggedright\arraybackslash}p{0.8cm} >{\raggedright\arraybackslash}p{3.0cm} X}
D-3 & \textbf{Assurance} & Grounds for sufficient confidence that an AI system (D-1), while operating within its defined scope, will achieve its intended outcomes without introducing unacceptable risks, throughout its lifecycle. 
\end{tabularx}

\begin{tabularx}{16cm}{>{\raggedright\arraybackslash}p{0.8cm} >{\raggedright\arraybackslash}p{3.0cm} X}
D-4& \textbf{Assurance Case} & A structured argument, supported by a body of evidence, demonstrating that an assurance claim (D-5) is true.
\end{tabularx}

\begin{tabularx}{16cm}{>{\raggedright\arraybackslash}p{0.8cm} >{\raggedright\arraybackslash}p{3.0cm} X}
D-5& \textbf{Assurance Claim } & A statement that relevant stakeholders must have sufficient confidence in for the assurance (D-3) of an AI system (D-1) to be established. 

NOTE: The top-level assurance claim of an AI system is the claim: “While operating within its defined scope, the system will achieve its intended outcomes without introducing unacceptable risks, throughout its lifecycle.” 

NOTE: The assurance case (D-4) for an AI system’s top-level assurance claim is used to establish the system’s assurance, i.e., the grounds for sufficient confidence that the system, while operating within its defined scope, will achieve its intended outcomes without introducing unacceptable risks, throughout its lifecycle.
\end{tabularx}

\begin{tabularx}{16cm}{>{\raggedright\arraybackslash}p{0.8cm} >{\raggedright\arraybackslash}p{3.0cm} X}
D-6 & \textbf{Assurance Plan} & An artifact which documents the technical and management approach for establishing system assurance and constructing a case for its top-level assurance claim.
\end{tabularx}

\begin{tabularx}{16cm}{>{\raggedright\arraybackslash}p{0.8cm} >{\raggedright\arraybackslash}p{3.0cm} X}
D-7 & \textbf{Assured} & A state in which relevant stakeholders have sufficient confidence in an AI system’s (D-3) top-level assurance claim (D-5) and have accepted its top-level assurance case (D-4). 
\end{tabularx}

\begin{tabularx}{16cm}{>{\raggedright\arraybackslash}p{0.8cm} >{\raggedright\arraybackslash}p{3.0cm} X}
D-8 & \textbf{Effectiveness} & The extent to which the goals of the system are attained, or the degree to which a system can be expected to achieve a set of specific mission requirements.
\end{tabularx}

\begin{tabularx}{16cm}{>{\raggedright\arraybackslash}p{0.8cm} >{\raggedright\arraybackslash}p{3.0cm} X}
D-9 & \textbf{Harm} & Undesirable impact to individuals, property, environment, society, or mission outcomes. \\
D-10 & \textbf{Hazard} & A real or potential condition that can lead to a harm (D-9). [Source: MIL-STD-882E\cite{mil-std-882} modified – list of harms replaced with “harm”].
\end{tabularx}

\begin{tabularx}{16cm}{>{\raggedright\arraybackslash}p{0.8cm} >{\raggedright\arraybackslash}p{3.0cm} X}
D-11 & \textbf{Lifecycle} & All phases of the system’s life, including design, research, development, test and evaluation, production, deployment (inventory), operations and support, and disposal. [Source: MIL-STD-882E\cite{mil-std-882}]
\end{tabularx}

\begin{tabularx}{16cm}{>{\raggedright\arraybackslash}p{0.8cm} >{\raggedright\arraybackslash}p{3.0cm} X}
D-12 & \textbf{Mishap} & An event or series of events resulting in a harm (D-9). [Source: MIL-STD-882E\cite{mil-std-882}, modified – list of harms replaced with word “harm”].\end{tabularx}

\begin{tabularx}{16cm}{>{\raggedright\arraybackslash}p{0.8cm} >{\raggedright\arraybackslash}p{3.0cm} X}
D-13 & \textbf{Risk} & A combination of the severity of the mishap (D-12) and the probability that the mishap (D-12) will occur. [Source: MIL-STD-882E\cite{mil-std-882}]  
\end{tabularx}


\section{Context}

There exists extensive literature on safety, risk assessments, and risk management that may be relevant to AI assurance. As defined in Section \ref{sec:definitions}, risk management is a major component of AI assurance, and those same concepts should be applied to ensuring effectiveness of the assured system. Our framework for assurance of AI systems is informed specifically by many of the publications on risk management, safety management, and AI assurance written by the United States (U.S.) National Institute of Standards and Technology (NIST), the International Organization for Standardization (ISO), International Electrotechnical Commission (IEC), U.S. Department of Defense, and the MITRE Corporation. 

NIST AI RMF provides a general framework for risk management of AI \cite{nist-ai-rmf}. MITRE presents a process for assurance of AIES, but without specific and detailed alignment to DOD processes, missions, and needs, because it is designed to be applicable to more application areas than DOD \cite{mitre-assurance}. Many of the concepts of our AI assurance framework are aligned with MITRE's process and concepts, including emphasis on risk management, various definitions, and the assurance plan. Our framework includes concepts relating to DOD-specific terms and processes, such as approvals needed for fielding systems.
ISO 31000 presents methods for management of organizational risk as well as concepts that are generally relevant to risk management \cite{iso-31000}. NIST SP 800-39 provides a guide for managing information security risk \cite{nist-800-39}. NIST SP 800-30 covers methods for conducting risk assessments, with focus on cyber security \cite{nist-800-30}. ISO/IEC 31010 provides guidance on the selection of risk assessment techniques to help an organization identify potential options and select one appropriate to the application and scenario. It additionally provides some guidance on application of some of the risk assessment techniques \cite{iec-31010}. ISO 14971 provides guidance for risk management of medical devices, including software as a medical device \cite{iso-14971}. Companion standards, such as BS/AAMI 34971:2023 provide guidance on the application of ISO 14971 to systems using AI machine learning \cite{bs-34971}. ISO 12100 provides general principles principles and methodology, including risk assessment and reduction techniques, for achieving safety for machinery\cite{iso-12100}. MIL-STD-882E provides detailed system safety standard practices for the DOD \cite{mil-std-882}. Other standards we made use of were those related to safety, such as IEC 61508 for functional safety \cite{iec-61508}, IEC 62061 for functional safety of machinery \cite{iec-62061}, and ISO IEC 82304 and ISO IEC 62304 for product standards, including safety, for medical device software \cite{iec-82304, iec-62304}.

Although much of this literature can be applied to AIES, it is often ignored for AI systems, and much of the literature is not designed to address risks that do not cause direct physical harm to people. Additionally, it is not clear how the myriad components, processes, requirements, etc., should be incorporated into DOD development and acquisition efforts. Our AI assurance framework deliberately leaves many details open to allow application of these standards in an integrated way. For example, it requires the use of risk assessment techniques, but leaves selection of appropriate methods to the developer. Our process attempts enable integration of many standards with relevant DOD MIL-STD documents, typical DOD development processes, and DOD acquisition processes, to inform how assured AI might be achievable.

Our process is also in alignment with established engineering design and product development strategies, much of which is also required by safety standards \cite{pahl2006engineering, eppinger1995product}.
These strategies call for identification of product requirements associated, engagement with key stakeholders, and performing iterations of design, implementation, and evaluation. Through these strategies, developers can sufficiently understand the problem and solution spaces, and provide sufficient evidence to stakeholders to establish assurance (D-3).

Although similar in intent to recognized standards like NIST SP 800-160\cite{nist-800-160} and ISO/IEC/IEEE 15026-1\cite{iso-15026-1}, the framework remains flexible enough to overlay both traditional milestone-based acquisitions and cutting-edge “campaigns of learning” approaches. It assumes that existing DOD artifacts for safety, cybersecurity, or ethics can be remapped into an assurance-case argument, thus avoiding redundant T\&E and fostering cross-domain synergy. However, the framework also identifies additional AI-focused documentation—such as adversarial ML assessments or generative AI bounding strategies—where needed to address emergent risks.

\section{Fundamental Concepts}

In this section we describe fundamental concepts to the AI assurance framework.

\subsection{A Risk-informed Approach}

Central to the success of such a risk-informed assurance framework are three guiding principles. The first is an emphasis on “unacceptable risk,” a notion that underscores how no system can be established as entirely risk-free, especially one using components as potentially complex and unpredictable as AI. Instead, assurance is evaluated on whether the known residual risks and the risks related to the understanding of risks, meet operational requirements, policy constraints, and risk tolerance. The second principle is transparency: stakeholders, from testers and operators to high-level decision-makers, must be able to see how the AI system’s claims are justified and which data or analyses were used to back them. This transparency is particularly important in an environment where public and international scrutiny may arise over AI’s role in decisions with moral or strategic implications. The final principle is continuous oversight. An AI-enabled performance is likely to change after initial deployment, either due to updates and learning by the AI or by changes to the scenarios. AI algorithms are often surprisingly brittle, failing in cases and modes that can be unexpected. Periodic or event-driven re-validation is essential to ensure continued performance of the system; the declaration of assurance for an AIES must be dependent on activities to maintain assurance.

By uniting these principles, unacceptable risk management, transparency in argumentation, and continuous oversight, the Department can more confidently accelerate AI fielding and achieve desired outcomes, without jeopardizing mission outcomes, public trust, or legal and ethical mandates. 

\subsection{AI-Specific Assurance Challenges}

Ensuring robust AI performance in defense applications involves unique technical and organizational challenges beyond those encountered with traditional software. Key AI-specific assurance challenges include the following:

\subsubsection{Data Bias and Quality}
AI models and a tester's ability to understand the AIES's performance are often heavily dependent on data. Data-driven AI model training may be limited by its training data. Biased, unrepresentative, or otherwise poor datasets can lead to undesirable outcomes. Testing is often be limited by test data, because data needs can exceed what is feasible to collect. Inability to test all foreseeable use cases, conditions, and uncommon scenarios, may mean that testers cannot complete sufficient evaluations. Ensuring data diversity, accuracy, and relevance is critical to avoid AIES performance degradation or improper test and assurance results.
For military AI, vetting data is especially important given the high stakes (e.g., targeting, intelligence) and the potential consequences of systematic errors.

\subsubsection{Model Complexity and Unpredictability}
Modern machine learning models (e.g., deep neural networks) often behave as ``black boxes,'' making it difficult to predict their behavior in all scenarios. Small input perturbations may produce chaotic changes in output, and there is typically no formal theory to guarantee behavior. This unpredictability complicates verification—unlike deterministic software; one cannot exhaustively test all conditions or easily reason about correctness. Ensuring reliability requires extensive validation across a wide range of scenarios and edge cases, including adversarial conditions.

\subsubsection{Lack of Explainability}
The opacity of AI decision-making processes poses a significant challenge for trust and oversight. When an AI system’s rationale cannot be clearly explained, operators and commanders may be unable or reluctant to trust its recommendations. Explainability can be considered a method to control some risks, providing users an ability to avoid certain mishaps (D-12). Moreover, lack of explainability hinders effective diagnosis of failures. Explainable AI techniques or assurance arguments (e.g., interpretable models, post-hoc explanations) may aid humans in understanding the justification for AI behavior in maintaining confidence in AI outputs.

\subsubsection{Dynamic Learning and Adaptation}
AI systems are frequently designed to learn and evolve over time, either through periodic retraining on new data or through online learning. Although this adaptability can enhance performance, it also means the system can change after deployment. Models may drift from their original behavior as the operational environment or input data distributions shift. Ensuring continued validity of an evolving AI system is a significant challenge—it requires ongoing monitoring, periodic re-testing, and careful configuration control to prevent regressions or unintended behaviors from emerging over time.

\subsubsection{Adversarial Threats and Robustness}
AI models are particularly susceptible to adversarial inputs and perturbations specifically crafted to deceive them. For instance, an image classifier might be fooled by subtle pixel manipulations, or a military decision aid might receive deceptive data intentionally provided by an adversary\cite{mitre-atlas}. These threat vectors (e.g., data poisoning during training, adversarial examples at runtime) necessitate rigorous red-teaming and robustness testing. Building resilience against such manipulations and detecting when the AI may be compromised is essential for mission-critical systems. Adversarial threats can be considered a external cause of risk to be considered during risk identification and assessment activities.

\subsubsection{Integration and Operational Constraints}
Deploying AI within complex military systems can introduce unique integration challenges. Interoperability with legacy systems, human operators, and under real-world conditions that may significantly differ from training simulations can be challenging. Factors such as real-time performance requirements, stringent safety constraints, and rigorous rules of engagement add additional layers of assurance complexity. Verifying that the AI will function appropriately under extreme or unpredictable battlefield conditions, and ensuring it fails safely if required, is a critical challenge extending beyond many conventional software testing practices.

\subsection{A Need for Consistency and a Unified Framework}


Although the DOD has robust processes for performance testing, safety evaluation, cybersecurity authorization, and ethics reviews, these traditionally operate in isolation, with each functional domain focused on specific criteria or checklists. AI-enabled systems cut across these domains, often exposing cross-domain vulnerabilities that stovepiped methods fail to anticipate. 

One illustrative scenario can arise with adversarial exploits. A penetration testing team may confine its scope to familiar application-layer vulnerabilities without examining how subtly perturbed input data can force an AI model into critical misclassifications or erroneous decisions. Meanwhile, a safety board might confirm that a robot’s mechanical subsystems meet reliability metrics but not realize how an Adversarial Machine Learning (AML) input (e.g., adversarial patches or malicious sensor feeds) could cause unsafe maneuvers in real-world environments \cite{kurakin2018adversarial}. As a result, the system viewed through the lens of siloed evaluations can appear acceptable in each narrow domain and underlying vulnerabilities could remain unrecognized. 

Inconsistent vocabularies and metrics across functional units exacerbate the difficulty. Cyber specialists measure risk through vulnerability scanning reports, threat modeling, and penetration test findings, often referencing frameworks such as NIST SP 800-53\cite{nist-800-53} for security controls. Safety engineers rely on hazard analyses and risk acceptance matrices like those in MIL-STD-882E\cite{mil-std-882}, and may talk about “hazards” and “mishaps” rather than “threats” or “exploits.” Ethics or legal offices might use entirely different terminologies, emphasizing compliance with directives like DODD 3000.09\cite{directive-3000-09} or concerns about algorithmic bias without referring to the same risk constructs or quantitative benchmarks. Commercial and academic developers might reference ISO and IEC standards like ISO 31000 and IEC 31010\cite{iso-31000, iec-31010}, or might have no background in standards. Simple terms like ``AI'' and ``model'' can hold very different meanings in different fields. These conceptual barriers make it challenging to coordinate efforts
into a cohesive understanding of risk.

Besides these gaps, stovepiped methods tend to be inefficient. Repeated testing, fragmented documentation, and disjointed sign-off processes prolong timelines and risk mitigation. Each domain’s separate review cycle can force AI developers to re-run tests or produce multiple, slightly varied reports to satisfy parallel offices. In a technology area such as AI, speed in development and deployment is paramount. To address rapidly evolving needs 
and to respond to the rapid evolution of AI algorithms, such inefficiencies should be avoided.
\section{The AI Assurance Process}


\begin{figure}[t]
    \centering
    \includegraphics[width=\textwidth]{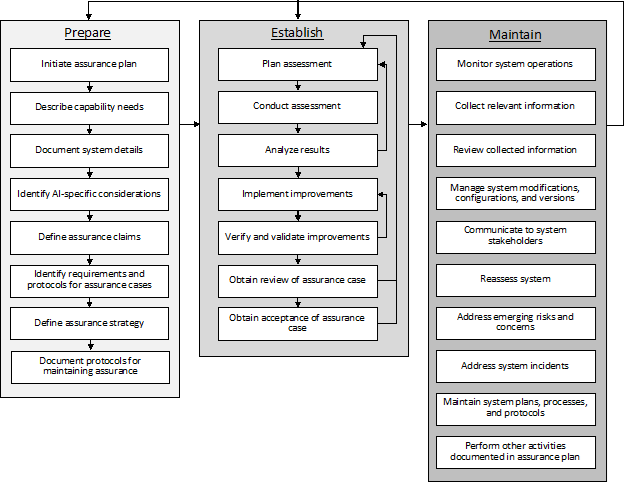}
    \caption{The phases and steps of the AI Assurance Process.}
    \label{fig:process}
    \vspace{-0pt}
\end{figure}


The AI assurance framework's process consists of three phases: (i) Prepare for assurance; (ii) Establish assurance; and (iii) Maintain assurance. Each phase is divided into a series of steps. This paper describes each step only at a high level. The process is presented in a linear order for clarity and readability. However, iteration among the different phases, steps, and tasks of the process is expected. Figure \ref{fig:process} illustrates the phases and steps of the process, as well as the paths for iteration among them. 

The AI assurance process is intended to integrate with existing DOD processes. It is not intended to replace any existing process for the development, fielding, or maintenance of DOD systems. Existing acquisitions, systems safety, cybersecurity, or test and evaluation processes, for example, may be leveraged to fulfill requirements of the AI assurance process. In particular, the artifacts and evidence from these processes may be integrated into the holistic case for the system’s assurance. System developers implementing the AI assurance process are encouraged to overlay it with their organizations existing processes in order to avoid duplication of effort. In the case where existing processes are inadequate, this framework may assist in modifying existing processes or establishing new ones.  

The process requires system developers to establish the protocols and criteria by which stakeholders may express their level of confidence in the effectiveness of the AI system and the acceptability of its associated risks. It does not prescribe any system-specific criteria or requirements which must be attained, such as requirements on acceptable levels of system performance or risk. 

The process does not prescribe specific methodologies for many of its activities, such as risk assessments and system evaluations. It also does not prescribe specific roles or responsibilities for these activities or for the other steps of the process. System developers implementing the process should seek guidance from their organization on the requirements and methodologies for conducting these activities as appropriate for the system and its purpose.  

\subsection{Prepare for Assurance}


For a given system, the activities necessary for achieving assurance must be tailored to the system and its context, including its purpose, operational environment, and potential impacts. 
The first phase of the assurance process is to prepare for assurance activities by documenting system information, defining assurance for the system and its context, and establishing a strategy to achieve this definition.  
Central to this phase is the specification of the system's assurance claims. These claims constitute the system's assurance – they define the qualities and criteria that relevant system stakeholders must have sufficient confidence in for the system to be assured.  

Some of the information required within this phase may be already documented in existing artifacts. Where this is the case, references to these artifacts may suffice for the satisfaction of the associated steps.  Additionally, some of the information documented within this phase such as the system’s operational environment, components, or supporting processes, may evolve over the course of the system’s development and sustainment. This phase provides an initial capture of this information, which may be iteratively updated and refined in the following phases. Iteration among steps within this phase may also be necessary. When possible, preparations for assurance activities should be included in early system development phases. 

The inputs to this phase are a capability need and the corresponding AI system or proposed AI system designed to meet the need, along with its development process documentation. Outputs of this phase are a set of documentation on the system, a set of assurance claims, and a strategy for achieving assurance for the system. 

The Prepare for Assurance phase consists of the following steps: 

\begin{enumerate}
\item \textbf{Initiate assurance plan}: Initialize the assurance plan artifact which will organize and contain the plans for establishing and maintaining assurance. This step includes creating documents, populating appropriate metadata, updating with appropriate document tracking systems, as well as establishing initial objectives, resources, and timeline for conducting assurance activities tailored to the AI system.
\item \textbf{Describe capability needs:} Provide a high-level description of the specific mission needs which inform the system’s development, including the operational environment, the current state, particular needs to be addressed, and the desired end state.
\item \textbf{Document system details:} Provide detailed documentation about the system, including a system description, mission description, system employment, details of expected system use, the scope of operations (including foreseeable misuse), system requirements, system architecture, and system stakeholders.
\item \textbf{Identify AI-specific considerations:} Create detailed documentation of considerations that include AI-specific policies, requirements, and guidance. Identify the AI components of the system including details on algorithms, models, architecture, data use, and rationale for the design. Identify the interactions between AI, humans, other systems, potential effects, and planned accountability for the system.
\item \textbf{Define assurance claims:} 
Define assurance claims and sub-claims that achieve the top-level assurance claim defined as, “While operating within its defined scope, the system will achieve its intended outcomes without introducing unacceptable risks, throughout its lifecycle.”
\item \textbf{Identify requirements and protocols for assurance cases:} Identify the types of evidence, validation methods, and documentation standards needed to support assurance claims. Additionally identify the methods for review and acceptance of assurance cases.
\item \textbf{Define assurance plan:} Lay out the content of the assurance plan, identifying assurance activities as well as roles, responsibilities, schedules, and resources for completing assurance activities.
\item \textbf{Document protocols for maintaining assurance:} 
Create the content of the assurance plan on protocols for maintaining assurance, which includes protocols for system monitoring, data collection, data review, managing system modification, managing system configurations, communicating with system stakeholders, assurance reassessments, managing emerging risks, managing incidents, maintaining system plans, and maintaining assurance protocols. 

\end{enumerate}

\subsection{Establish Assurance}


The second phase of the assurance process is to establish the system’s assurance through an iterative series of assessments and improvements. 
This phase is driven by assessments, which may be employed throughout the system’s development to identify issues impacting system assurance, produce evidence for the system’s assurance cases, and produce other information to inform development efforts.

Earlier assessments within the development process are often exploratory in nature and may be used to identify potential hazards, inform system designs, or test component functionality. Later assessments may be used to evaluate system safety or validate system performance within an operationally realistic environment. 
Throughout the phase, information from assessments and other activities are used as evidence to incrementally build the system’s assurance cases. A system advances from the phase after the acceptance of its top-level assurance case. 

NOTE: Within this framework, the term “assessment” is used in a broad sense to include a variety of systematic activities aimed at determining the nature of something, including tests, experiments, surveys, analyses, audits, inspections, and simulations. 

The inputs to this phase are an AI system, its development process documentation, and an assurance plan containing the documentation from the previous phase. The outputs of this phase are an assured AI system, an approved assurance plan, and a set of assurance cases that substantiate the system’s assurance claims.  

The Establish Assurance phase consists of the following steps: 

\begin{enumerate}
\item \textbf{Plan assessment:} Define the purpose, scope, requirements, guidance, methods, resources, and design of the planned iteration of assessment. Assessment design may differ depending on risks being evaluated, nearness to fielding, and preferences of the assessors. 
\item \textbf{Conduct assessment:} Perform the planned assessment, documenting observations.
\item \textbf{Analyze results:} Analyze the results of the assessment according to the methodologies and procedures selecting during assessment design. Determine issues and results relevant to system assurance to identify gaps, issues, or risks that must be addressed or further assessed.
\item \textbf{Implement improvements:} 
For each issue impacting assurance, identify potential improvements, consider the feasibility and value of improvements, and implement chosen improvements.
\item \textbf{Verify and validate improvements:} 
Verify the implementation of the improvements and validate the effectiveness of the improvement in addressing the corresponding issue impacting system assurance. This step may essentially iterate on the assessment on the system after implementation of improvements.
\item \textbf{Obtain review of assurance case:} 
Obtain a review of the assurance case according to the protocols documented during the Prepare for Assurance phase. Review should be performed by appropriate stakeholders at appropriate times, as defined in the assurance plan.
\item \textbf{Obtain acceptance of assurance case:} 
Obtain approval of the assurance case according to the protocols documented during the Prepare for Assurance phase. For the system to be assured, the system's top-level assurance claim must be accepted. Continue to iterate through this phase until that acceptance is obtained.

\end{enumerate}

NOTE: The steps of this phase are designed to be performed iteratively, with each iteration scoped to support the current development objectives and level of system maturity. For example, multiple iterations of plan assessment, conduct assessment, and analyze results will likely need to be performed over the course of system development.  

NOTE: While the steps are presented in a sequential manner, iteration among steps is also necessary. For example, the results from an initial assessment may require a follow-on assessment to confirm its findings. Figure~\ref{fig:process} illustrates the steps of the phase and their iterative nature.

\subsection{Maintain Assurance}


After an assured system has been fielded, its assurance is contingent upon its continued operation within scope, its continued effectiveness, and the continued acceptability of its risks. 
The third phase of the assurance process is to maintain the system’s assurance through the protocols documented within the assurance plan. 

These protocols maintain assurance by providing critical system information and, as appropriate, managing system incidents, modifications, and risks. This phase continues throughout system operations. During this phase, various activities or findings can trigger a return to other phases of the assurance process. 

NOTE: The activities of this phase are not designed to be performed in sequence; rather, they have different mechanisms and occasions for their implementation. Some may be implemented continuously, others periodically, and others only when triggered by other events. Details of when and how they should be implemented should be documented in the assurance plan.  

The Maintain Assurance phase includes the following activities: 

\begin{itemize}
\item \textbf{Monitor system operations:} Monitor system operations to detect anomalies and track key parameters of system performance. Some notable parameters for system monitoring may include measures algorithm performance, parameters of the operational environment, indicators of data drift, indicators of system health, and indicators of mishaps.  
\item \textbf{Collect relevant information:} 
Separate from direct monitoring of the system, collect other relevant data such as feedback from operators, feedback from other stakeholders, new regulations relevant to the system, and information about system deployments, maintenance, and failures.
\item \textbf{Review collected information:} 
Collected information must be reviewed with some frequency to identify risks and recognize any needs for revisions to the assurance plan. 
\item \textbf{Manage system modifications, configurations, and versions:} 
System modifications, configurations, and versions must be managed appropriately to ensure that the system's assurance is maintained and that only the correct configurations of the system, its components, and its software are used across deployments. Protocols for management must include information requirements for modification requests, processes to submit and review modification requests, processes for managing system configurations, and processes for verifying system configurations. Model re-training may be a common form of modification covered within this protocol.
\item \textbf{Communicate to system stakeholders:} 
Relevant system information such as status, modifications, decisions, and incidents must be communicated to appropriate stakeholders throughout the system's lifecycle.
\item \textbf{Reassess system:} 
The system must be reassessed at regular intervals, after major system changes, and after identification of potential unanticipated risks. Protocols on reassessment should describe the types of reassessments, the frequency, triggering events, and processes for reassessment. The process may look similar to assessments performed in the Establish Assurance phase.
\item \textbf{Address emerging risks and concerns:} 
Emerging risks and concerns that are identified must be addressed in order to maintain assurance. 
\item \textbf{Address system incidents:} 
Incidents, such as system mishaps, must be addressed promptly and effectively in order to understand causes, mitigate harms, and prevent future occurrences.
\item \textbf{Maintain system plans, processes, and protocols:} 
The system's plans, processes, and protocols, including those for maintaining assurance, must be maintained throughout the system lifecycle.
\item \textbf{Perform other activities documented in assurance plan:} 
Throughout the assurance process, other activities to maintain system assurance may be identified and documented in the assurance plan. For example, these might include periodic model re-training, system maintenance, or operator training. These must be performed as documented.

\end{itemize}


\section{Conclusion}

This paper has presented the basics of a claims-based AI assurance framework designed to balance rapid fielding
with sufficiently rigorous oversight. By integrating safety, security, performance, and ethical or legal mandates
into a single argument-based approach, the framework consolidates artifacts and analyses that might otherwise
remain stovepiped. This comprehensive perspective helps identify AI-specific vulnerabilities, such as data
drift, emergent behaviors, and compliance gaps, early in the development lifecycle. However, more detail on the
steps of the process and alignments of the process with existing policies and processes is necessary for practical
application of the process. 
Programs can benefit from a clearer understanding of relevant risks and more efficient coordination across
different functional offices. As a result, this assurance methodology promotes trust, accountability, and adaptability, ensuring the DOD can capitalize on AI’s strategic potential without compromising on mission-critical
standards of effectiveness and safety.

\subsection*{Acknowledgments}

This technical data deliverable was developed in part using contract funds under Basic Contract No. W56KGU-18-D-0004.


\newpage

\bibliography{report} 
\bibliographystyle{spiebib} 

\end{document}